\documentclass[letterpaper, 10 pt, conference]{ieeeconf}  
\usepackage{amsmath}
\usepackage{amssymb}
\usepackage{nccmath}
\usepackage{mathtools}
    {\end{bmatrix}\end{medsize}}%
\usepackage[linesnumbered,ruled,vlined]{algorithm2e}
\usepackage{algorithmic}
\usepackage{graphicx}
\usepackage{graphics}
\usepackage{tikz}
\usetikzlibrary{calc,patterns,decorations.pathmorphing,decorations.markings}
\usetikzlibrary{positioning}
\usetikzlibrary{automata,positioning}
\makeatletter
\def\BState{\State\hskip-\ALG@thistlm}
\makeatother 
\graphicspath{{images/}}
\usepackage{textcomp}
\usepackage{booktabs}
\usepackage{amsmath,amssymb}
\usepackage{lipsum} 
\usepackage{graphicx}
\usepackage{indentfirst}
\usepackage{color,soul}

\IEEEoverridecommandlockouts                              
\usepackage{amsmath} 

\title{\LARGE \bf
Toward Achieving Formal Guarantees for Human-Aware Controllers in Human-Robot Interactions 
}

\author{Rachel Schlossman$^{1}$, Minkyu Kim, Ufuk Topcu, Luis Sentis 
\thanks{*This work was supported by NASA Space Technology Research Fellowship 80NSSC17K0188.}
\thanks{$^{1}${\tt\small rachel.schlossman@utexas.edu}}%
\thanks{Authors are with The Departments of Mechanical Engineering (R.S., M.K.) or Aerospace Engineering (U.T., L.S.), University of Texas at Austin, Austin, TX 78712-0292, USA}%
}

\begin{document}

\maketitle
\thispagestyle{empty}
\pagestyle{empty}


\begin{abstract}
With the primary objective of human-robot interaction being to support humans' goals, there exists a need to formally synthesize robot controllers that can provide the desired service. Synthesis techniques have the benefit of providing formal guarantees for specification satisfaction. There is potential to apply these techniques for devising robot controllers whose specifications are coupled with human needs. This paper explores the use of formal methods to construct human-aware robot controllers to support the productivity requirements of humans. We tackle these types of scenarios via human workload-informed models and reactive synthesis. This strategy allows us to synthesize controllers that fulfill formal  specifications that are expressed as linear temporal logic formulas. We present a case study in which we reason about a work delivery and pickup task such that the robot increases worker productivity, but not stress induced by high work backlog. We demonstrate our controller using the Toyota HSR, a mobile manipulator robot. The results demonstrate the realization of a robust robot controller that is guaranteed to properly reason and react in collaborative tasks with human partners. 
\end{abstract}



\section{INTRODUCTION}

As robots become more embedded into our everyday lives and begin to collaborate with humans, a large potential emerges to boost human productivity by eliminating unnecessary human chores in workplaces \cite{sutherland2007seven}. This potential can only be realized by robot control systems that process and react to human needs. A survey of human-aware robot navigation shows that many researchers have studied motion and task plan generation for socially-aware robots, for activities ranging from operating in human-occupied areas to engaging in social cues \cite{kruse2013human}. However, these methods often lack robustness to disturbances and/or formal guarantees of goal achievement. Formal methods have been leveraged in robotic applications, but there is a growing need to apply these techniques to robots that continuously and directly interact with humans. Our work takes a step toward addressing this need.

\begin{figure}[tp]
\def\svgwidth{.98\columnwidth}
\begin{center}
\begingroup%
  \makeatletter%
  \providecommand\color[2][]{%
    \errmessage{(Inkscape) Color is used for the text in Inkscape, but the package 'color.sty' is not loaded}%
    \renewcommand\color[2][]{}%
  }%
  \providecommand\transparent[1]{%
    \errmessage{(Inkscape) Transparency is used (non-zero) for the text in Inkscape, but the package 'transparent.sty' is not loaded}%
    \renewcommand\transparent[1]{}%
  }%
  \providecommand\rotatebox[2]{#2}%
  \newcommand*\fsize{\dimexpr\f@size pt\relax}%
  \newcommand*\lineheight[1]{\fontsize{\fsize}{#1\fsize}\selectfont}%
  \ifx\svgwidth\undefined%
    \setlength{\unitlength}{202.5bp}%
    \ifx\svgscale\undefined%
      \relax%
    \else%
      \setlength{\unitlength}{\unitlength * \real{\svgscale}}%
    \fi%
  \else%
    \setlength{\unitlength}{\svgwidth}%
  \fi%
  \global\let\svgwidth\undefined%
  \global\let\svgscale\undefined%
  \makeatother%
  \begin{picture}(1,0.48148148)%
    \lineheight{1}%
    \setlength\tabcolsep{0pt}%
    \put(0,0){\includegraphics[width=\unitlength,page=1]{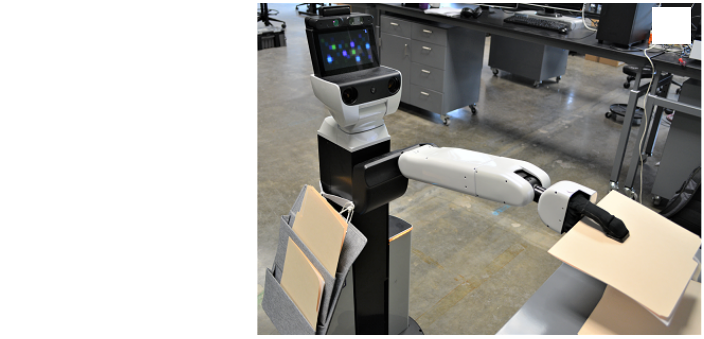}}%
    \put(0.93230026,0.43809653){\color[rgb]{0,0,0}\makebox(0,0)[lt]{\lineheight{1.25}\smash{\begin{tabular}[t]{l}(b)\end{tabular}}}}%
    \put(0,0){\includegraphics[width=\unitlength,page=2]{pickup_dropoff.pdf}}%
    \put(0.01882263,0.43845109){\color[rgb]{0,0,0}\makebox(0,0)[lt]{\lineheight{1.25}\smash{\begin{tabular}[t]{l}(a)\end{tabular}}}}%
  \end{picture}%
\endgroup%

  \caption{(a) The Toyota Human Support Robot, which we use as our experimental platform for human-informed work delivery, picking up completed work. (b) The HSR dropping off completed work at the Inventory Station.}  \label{fig:pickup_dropoff}
    \vspace{-20pt}  
\end{center}
\end{figure}

Reactive synthesis has been applied in contexts where disturbances are unavoidable, such as the DARPA Robotics Challenge \cite{maniatopoulos2016reactive} and other challenging setups \cite{he2017reactive}, to provide formal guarantees for specification realizability. There is a gap in robotics and formal method literature with respect to applying these methods to direct human-robot interaction (HRI). In reactive synthesis problems, humans are often framed as randomly or periodically interfering with the robot's goal \cite{he2017reactive}, \cite{zhao2016high}. In \cite{wang2016task}, a reactive synthesis problem is formulated to generate a policy for a robot to reach a goal position in a simulated kitchen scenario, but the only human interaction involves avoiding two moving chefs. We seek to be robust to a larger variety of disturbances, and to generate policies in which humans and robots continuously interact.    

There is much interest in the HRI community for robot controllers to consider human factors to boost human productivity \cite{ramachandran2017give}, \cite{chen2018planning}. Several research groups are exploring formal verification methods for robots to interactively support humans \cite{wu2017toward}, but few groups have incorporated human factors into these methods. In \cite{webster2016toward}, the authors verify whether a robot assistant can reach commanded positions and deliver medicine. In addition to these types of interactions, we also explore additional knowledge of human requirements to improve collaborative task execution. Ref. \cite{kwiatkowska2017cognitive} takes a step in this direction: A cognitive model of trust is incorporated into a stochastic multi-player game and probabilistic rational temporal logic specifications are proposed, but probabilistic model checking is left as future work. In \cite{porfirio2018authoring}, social norms for a hand-off task are represented as transition systems, and model checking is performed to verify successful task completion. In contrast to this work, our framework uses reactive synthesis to prevent specification violations, while being robust to uncertainties. 



Although it is impossible to generalize human behavior, works like \cite{sadigh2016information}, \cite{rausch2018modeling} still demonstrate the insight to be gained by considering human models for decision-making. For example, in \cite{feng2016synthesis}, hypothetical human models that consider human proficiency and stress are employed to synthesize paths for semi-autonomous operation of drones with human operators. Similarly, we focus on studying the implications of incorporating human models. Our approach provides the flexibility to update the human model for effective and personalized human-robot interactions. 

The goal of this paper is to demonstrate a proof-of-concept for devising control policies via reactive synthesis that consider and improve human working behavior. In doing so, we generate a controller that
is robust to disturbances and provides formal guarantees for specification satisfaction in an HRI scenario. The main contribution of our work is a study on reactive synthesis that incorporates human factors for human-aware robot cooperative tasks. We consider an HRI case study in which we construct a model of human workers in a workplace as transition systems to devise a robot controller to deliver and pick up work while considering the human's needs. To this end, we formalize system specifications using linear temporal logic. We use reactive synthesis to automatically construct a controller that meets all system specifications and a human's productivity needs. We then demonstrate the reactive controller on the Toyota Human Support Robot (HSR) (Fig. \ref{fig:pickup_dropoff}). Ultimately, we explore the question of how robots can make humans more productive by limiting unnecessary human tasks

\section{PRELIMINARIES}
Our notation employs the formalisms of \cite{baier2008principles}, which are summarized below:

\textbf{Definition 1:} A transition system, $TS$, is a tuple $TS = (S, Act, \rightarrow, I, AP, L)$ where $S$ is a set of states, $Act$ is a set of actions, $\rightarrow\subseteq S \times Act \times S$ is a transition relation,  $I \subseteq S$ is a set of initial states, $AP$ is a set of 
(Boolean) atomic propositions, and $L : S \mapsto 2^{AP}$ is a labeling function.

\textbf{Definition 2:} An infinite path fragment, $\pi=s_0s_1s_2...$, for $s_i \in S$,  is an infinite sequence of states such that $s_{i+1} \in \{s_i' \in S : \exists\alpha \in  Act \mid s_i \xrightarrow[]{\alpha} s_i' \} \: \forall i \geq 0$. An infinite path fragment is a path if the initial state, $s_0 \in I$. The set of paths in $TS$ is denoted as $Paths(TS)$. 

\textbf{Definition 3:} The trace of $\pi$, $trace(\pi)=L(s_0)L(s_1)L(s_2)...$, is a sequence of sets of atomic propositions that are true in the states along the path. The set of traces of $TS$ is defined by $Traces(TS) = \{trace(\pi) : \pi \in Paths(TS)\}$  

\textbf{Definition 4:} A linear-time (LT) property, $P$, over atomic propositions in $AP$, is a set of infinite sequences over $2^{AP}$. $TS$ satisfies $P$, represented by $TS \models P$, iff $Traces(TS) \subseteq P$.  

\textbf{Definition 5:} Linear-temporal logic (LTL) is a formal language to represent LT properties. The operators used in this paper to construct LT formulas are conjunction ($\wedge$), disjunction ($\vee$), next ($\bigcirc$), eventually ($\lozenge$), globally ($\square$), implication ($\rightarrow$), and negation ($\lnot$).  Let $\Phi$ be an LTL formula over $AP$. $TS$ satisfies $\Phi$, represented by $TS \models \Phi$, iff $\pi \models \Phi$ for all $\pi \in Paths(TS)$.


\section{Work Delivery with Human Backlog Model}

This case study examines a robot operating in a work environment and is inspired by \cite{jorgensen2017exploring}. The robot drops off new work (``deliverables") at the Human Workstation, picks up completed work from the Human Workstation, and drops the completed work off at the Inventory Station. The robot's contributions thereby eliminate unnecessary movement by the human to pick up and drop off work. The goal is for the robot to operate with an awareness of the human's backlog, defined as the amount of uncompleted work at the Human Workstation. Assessing human backlog is an ongoing area of research \cite{heard2018survey}. We take backlog to be an indicator of stress levels, as too little work can cause boredom and too much work can result in higher levels of frustration \cite{macdonald2003impact}. We seek to synthesize a controller that guarantees, despite system disturbances, that the human always has work to complete and is not over-stressed by work demands.
 
\subsection{Modeling} 

\subsubsection{Human Model}\label{sec:delivery_human_model}

In this scenario, the human is always present in the Human Workstation. (Work breaks are addressed in Sec. \ref{sec:experiments}.) The human's backlog, $BL$, can range from $0\%$ to $100\%$, relative to the maximum amount of uncompleted work that can be present in the workstation. When the robot is not present in the Human Workstation, the human works whenenever there is uncompleted work. We consider a simple, discrete linear $BL$ model that is a function of $\Delta T$ time steps that each last $t_d$ seconds:

\begin{equation}\label{eqn:BL_model}
BL(\Delta T) = BL_{init} - \gamma \Delta T, \, \Delta T \in \{0, 1, 2,...\},
\end{equation}

\noindent where $BL_{init}$ is the initial amount of backlog at the Human Workstation, and $\gamma$ is the work reduction rate per $t_d$ seconds. The value of $BL_{init}$ is updated each time the robot comes to the Human Workstation to deliver work.  

We now formalize the way $BL$ may change between time steps. There is uncertainty in how much $BL$ decreases during each state transition, as the reduction value depends on how much time the robot requires to transition between states in the real-world execution. The worker's $BL$ can  decrease by integer multiples of $\gamma$ each time step. We assume that $BL$ may decrease by up to $5\gamma$, based on the maximum amount of time the robot requires for its most challenging manipulation task. We consider two possibilities for how $BL$ may shrink. When the robot is traveling between locations in the workspace, we define the formula $v_1$ such that the following holds:


\begin{equation}\label{eqn:wl_small_decrease}
\bigcirc v_1 \triangleq  \bigvee_{k=0}^{2} (BL-k\gamma).
\end{equation}


It may require more time for the robot to drop off completed work at the Inventory Station than to travel between locations, and so we allow for greater $BL$ reductions between time steps for this task:

\begin{equation}\label{eqn:wl_big_decrease}
\bigcirc v_2 \triangleq  \bigvee_{k=0}^{5} (BL-k\gamma).
\end{equation}

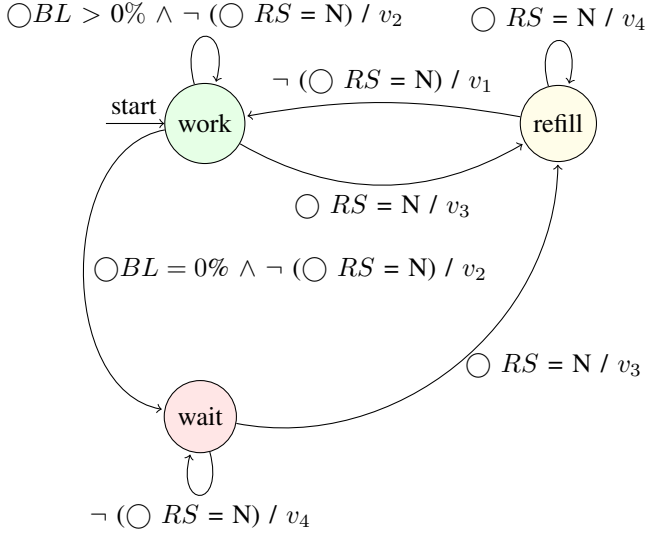
\begin{figure}[t]
\vspace{10pt}
\centering
\begin{tikzpicture}[shorten >=1pt,node distance=4.7cm,on grid,auto, scale=0.3]
   \node[state, fill={green!10}] (q_0)   {work}; 
   \node[state, fill={red!10}] (q_1) at (-0.2, -13) {wait}; 
   \node[state, fill={yellow!10}] (q_2) [right=of q_0] {refill}; 
      \draw[<-] (q_0) -- node[above] {start}
  ++(-4.5cm,0);
    \path[->] 
    (q_0) edge [below, bend right=80, right=0.3]  node {$\bigcirc BL = 0$\%  $\wedge$ $\lnot$ ($\bigcirc$ $RS$ = N) / $v_2$} (q_1)
          edge  [below, bend right] node  {$\bigcirc$ $RS$ = N / $v_3$} (q_2)
          edge [loop above] node {$\bigcirc BL$ $>$ 0$\%$ $\wedge$ $\lnot$ ($\bigcirc$ $RS$ = N) / $v_2$} ()
    (q_1) 
          edge [below, bend right = 50, right=0.3]  node {$\bigcirc$ $RS$ = N / $v_3$} (q_2)
          edge [loop below] node {$\lnot$ ($\bigcirc$ $RS$ = N) / $v_4$} ()
    (q_2) 
           edge  [above, bend right=10] node {$\lnot$ ($\bigcirc$ $RS$ = N) / $v_1$} (q_0)
          edge [loop above] node {$\bigcirc$ $RS$ = N / $v_4$} ();
\end{tikzpicture}
\vspace{-20pt}
\caption{Human Model with three states. Transitions are triggered by guards, $g$, and there are corresponding outputs, $y$. In this transition system, edges are labeled in a $g$ / $y$ format. In this system, the outputs are described by $v_2$, $v_3$, and $v_4$.}
\label{fig:human_model_delivery}
\vspace{-15pt}
\end{figure}

The human transitions from the ``work" state to the ``wait" state if the human has completed all of her work and the robot has not yet arrived to deliver more work. The human transitions to the ``refill" state when the robot is present in the Human Workstation and delivers more work. The robot being in the Human Workstation is equivalent to the robot state, $RS$, being equal to N. When the robot arrives at the Human Workstation, $BL$ grows by $\delta$. The human then returns to the working state when the robot departs from the workstation. As seen in Fig. \ref{fig:human_model_delivery}, there are guards based on $BL$ and robot behaviors which determine allowable state transitions. The $BL$ variable is tracked, and grows and shrinks according to $v_1$, $v_2$, $v_3$, and $v_4$. The formula $v_3$ captures $BL$ growth when the robot arrives at the Human Workstation:

\begin{equation}\label{eqn:work_delivery}
 \bigcirc v_3 \triangleq BL+\delta,    
\end{equation}

\noindent and $v_4$ captures when $BL$ does not change between time steps: 

\begin{equation} \label{eqn:BL_stays_constant}
\bigcirc v_4 \triangleq BL.    
\end{equation}


Fig. \ref{fig:human_model_delivery} demonstrates the possible non-determinism in how $BL$ changes with each time step, and planning for this uncertainty is discussed further in Sec. \ref{sec:reactive_synthesis}.

\subsubsection{Robot Model}

The robot moves in a 1D grid. There are N+1 grid spaces, with the grid spaces labeled as 0 through N. Space 0 corresponds to the Inventory Station, and (\ref{eqn:wl_big_decrease}) is valid in this location when the robot drops off completed work. As discussed in the previous section, space N is the Human Workstation. The robot's actions are Go$_{\text{Sj}}$, which indicate that the robot is currently moving to position j $\in$ 0,1,...,N in the grid. The robot is free to move within this grid, except for when there is an obstacle present in a grid space blocking a path. We define the atomic proposition, ``an obstacle is present in State j," as $\mathcal{O}_j \enspace \forall j$ = 1,2,...,N-1. The transition system is shown in Fig. \ref{fig:robot_model_delivery}.

\begin{figure}[t]
    \centering
\begin{tikzpicture}[shorten >=1pt,node distance=2.8cm,on grid,auto] 
   \node[state,initial, fill={gray!10}] (q_0)   {0}; 
   \node[state, fill={gray!10}] (q_1) [right=4.5 of q_0]  {1}; 
   \node[state, fill={gray!10}] (q_2) [above=of q_1]  {2}; 
   \node[state, fill={gray!10}] (q_5) [left=4.5 of q_2]  {3}; 
    \path[->] 
    (q_0) edge  [bend right, below]  node {$\lnot g_1$ / Go$_{\text{S1}}$} (q_1)
        edge [loop above] node { ($g_1 \vee \lnot g_1$) / Go$_{\text{S0}}$} ()
    (q_1) 
          edge [above]  node {($g_2 \vee \lnot g_2$) / Go$_{\text{S0}}$} (q_0)
          edge [loop below] node {($g_2 \vee \lnot g_2$) / Go$_{\text{S1}}$} ()
          edge []  node {$\lnot g_2$ / Go$_{\text{S2}}$} (q_2)
    (q_2) 
          edge [bend left]  node {$\lnot g_1$ / Go$_{\text{S1}}$} (q_1)
          edge [loop above] node {($g_1 \vee \lnot g_1$) / Go$_{\text{S2}}$} ()
          edge []  node {($g_1 \vee \lnot g_1$) / Go$_{\text{S3}}$} (q_5)    
     (q_5) 
          edge [bend left]  node {$\lnot g_2$ / Go$_{\text{S2}}$} (q_2)
          edge [loop above] node {$g_2$ / Go$_{\text{S3}}$} ();
\end{tikzpicture}   
\vspace{-20pt}
    \caption{Robot Model with N = 3. State 0 is the Inventory Station and State 3 is the Human Workstation. Each edge of the $TS$ is labeled with a guard and an action. The presence of an obstruction in one of the robot's adjacent positions can restrict the robot's next action. The guards express whether or not there is an obstacle blocking the robot from proceeding to a neighboring state, and we define $g_1 \triangleq \mathcal{O}_1$ and $g_2 \triangleq \mathcal{O}_2$} \label{fig:robot_model_delivery}
    \vspace{-15pt}
\end{figure}
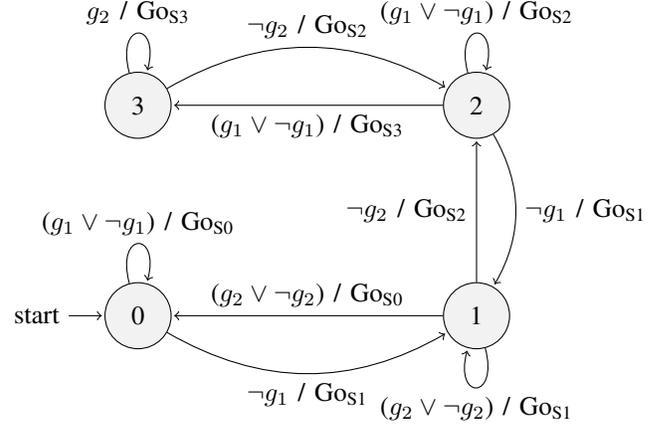

\begin{figure*}[ht]
\centering
\begin{tikzpicture}[shorten >=1pt,node distance=5.3cm,on grid,auto] 
   \node[state,initial, fill={green!10}] (q_0)   {$0_{work}$}; 
   \node[state, fill={red!10}] (q_1) at (0, -5)  {$0_{wait}$}; 
   \node[state, fill={green!10}] (q_2) [right=6.1 of q_0]  {$1_{ work}$}; 
   \node[state, fill={red!10}] (q_3) [right=6.1 of q_1]  {$1_{ wait}$}; 
   \node[state, fill={green!10}] (q_4) [right=6.1 of q_2]  {$2_{ work}$}; 
   \node[state, fill={red!10}] (q_5) [right=6.1 of q_3]  {$2_{ wait}$};  
  \node[state, fill={yellow!10}] (q_10) [below right=3.8 of q_4]  {$3_{refill}$};  
    \path[->] 
    (q_0) edge node [left] {$g_4$/Go$_{\text{S0}}$/$v_2$}         (q_1)
        edge [loop above] node {$g_3$/Go$_{\text{S0}}$/$v_2$} ()
        edge [bend left]  node {$g_3$/Go$_{\text{S1}}$/$v_1$} (q_2)
        edge [bend left=10, right=40] node {$g_4$/Go$_{\text{S1}}$/$v_1$} (q_3)
    (q_1) 
        edge [loop below] node {Go$_{\text{S0}}$/$v_4$} ()
        edge [below, bend left]  node {Go$_{\text{S1}}$/$v_4$} (q_3)        
    (q_2) edge [bend right = 20]  node [right] {$g_4$/Go$_{\text{S1}}$/$v_1$}         (q_3)
        edge [loop above] node {$g_3$/Go$_{\text{S1}}$/$v_1$} ()
        edge [above, bend left = 20]  node {$g_3$/Go$_{\text{S0}}$/$v_2$} (q_0)
        edge [bend left]  node {$g_3$/Go$_{\text{S2}}$/$v_1$} (q_4) 
        edge [bend left=10, right=1.5] node {$g_4$/Go$_{\text{S2}}$/$v_1$} (q_5)
        edge [bend right=5, left=40] node {$g_4$/Go$_{\text{S0}}$/$v_2$} (q_1)
    (q_3)
        edge [loop below] node {Go$_{\text{S1}}$/$v_4$} ()
        edge [below, bend left]  node {Go$_{\text{S2}}$/$v_4$} (q_5) 
        edge [bend left]  node {Go$_{\text{S0}}$/$v_4$} (q_1)
    (q_4) 
        edge [bend right = 20] node {$g_4$/Go$_{\text{S2}}$/$v_1$}         (q_5)
        edge [loop above] node {$g_3$/Go$_{\text{S2}}$/$v_1$} ()  
        edge [bend left=60]  node {Go$_{\text{S3}}$/$v_3$} (q_10)
        edge [above, bend left = 20]  node {$g_3$/Go$_{\text{S1}}$/$v_1$} (q_2)
        edge [bend right=10, left=1.5] node {$g_4$/Go$_{\text{S1}}$/$v_1$} (q_3)
    (q_5)
        edge [loop below] node {Go$_{\text{S2}}$/$v_4$} ()
        edge [bend right = 20]  node {Go$_{\text{S3}}$/$v_3$} (q_10)
        edge [bend left]  node {Go$_{\text{S2}}$/$v_4$} (q_3)
    (q_10)
        edge node [right] {Go$_{\text{S2}}$/$v_1$} (q_4)
        edge [loop below] node {Go$_{\text{S3}}$/$v_4$} (); 
\end{tikzpicture}
\caption{HRI Work Delivery and Pickup Transition System with N = 3. The edges of the transition system are labeled first by guards ($g_3$, $g_4$), then by actions, and lastly by pertinent output variables ($v_1$, $v_2$, $v_3$, $v_4$). The guards which determine whether the human is working or waiting are expressed by $g_3 \triangleq$ $\bigcirc BL >$ 0\% and $g_4 \triangleq$ $\bigcirc BL$ = 0\%. As illustrated in Fig. \ref{fig:robot_model_delivery}, the robot cannot move to a neighboring state if it contains an obstacle, but we do not show this guard due to space limitations.} \label{fig:combined_model_delivery}
\vspace{-15pt}
\end{figure*}
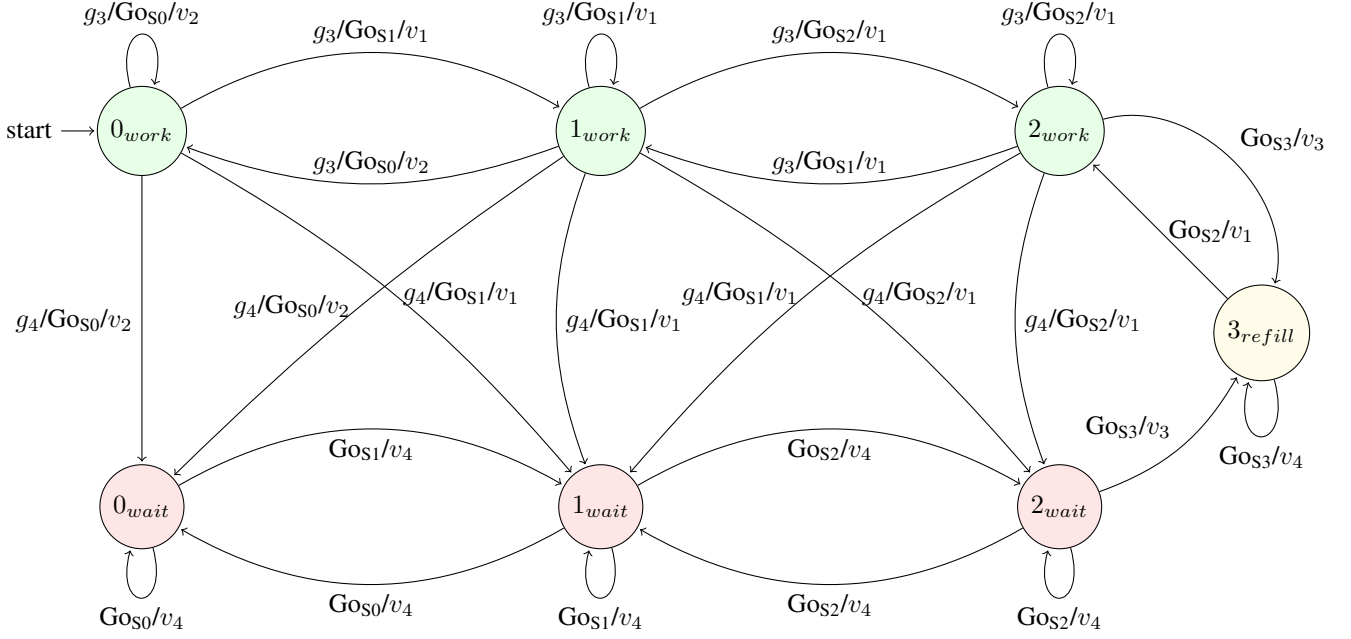

\subsubsection{Full HRI System Model}

A transition system is then formulated to capture that the robot drops off deliverables and picks up completed work at the Human Workstation, and drops off completed work at the Inventory Station. The robot also operates with an awareness of the human's work backlog, which will allow for controller synthesis that considers $BL$. To combine and synchronize the human and robot systems, the two separate human and robot models were used to create states which represent both the human and robot at each time step. 
The  transition system, shown in Fig. \ref{fig:combined_model_delivery}, is expressed as $TS_{HRI}$ = $(S_1, Act_1, \rightarrow, I_1, AP_1, L_1)$ where:
\begin{itemize}
  \item $S_1$ = $\{$$j_{work}$, $j_{wait}$, $N_{refill}$$\}$ $\forall$ $j$ = 0,1,...,N-1
  \item $Act_1$ = $\{$ Go$_{\text{Sj}}$ $\}$ $\forall$ $j$ = 0,1,...,N
  \item $I_1$ = $\{ 0_{work} \}$
   \item $L_1(0_{work})$ =  $L_1(0_{wait})$ = Robot is at Inventory Station.
    \item $L_1$($N_{refill})$ = Robot is at Human Workstation.    
\end{itemize}

\subsection{Reactive Synthesis} \label{sec:reactive_synthesis}
It is necessary to incorporate robustness to uncertainty in our approach in order for the robot to pick up completed work, drop off deliverables, and reason about the human's $BL$ in a real environment. We consider a two-player game in which the robot's actions are controllable.  Obstacle interference, success of dropping off completed work, and the backlog reduction rate act as the uncontrollable environment. The robot and the environment take turns executing actions, and we seek to automatically synthesize a robot controller strategy that allows a system specification to be realizable despite any antagonistic actions executed by the environment. To meet the system requirements while handling external disturbances, we formulate this scenario as a reactive synthesis problem in which plant actions are controllable and environment actions are uncontrollable. We seek to find a strategy that will uphold a specification  no matter how the environment selects its actions for all time \cite{piterman2006synthesis}.

To automatically synthesize a controller, we must formalize the specifications that describe the possible environment behaviors. It is possible in a work environment that people may pass through states S$_{d}$ = $\{$$j_{work}$, $j_{wait}$$\}$ $\forall j : 0 < j < N$ and obstruct the robot from proceeding into an adjacent position in the workspace. We impose as a safety specification on the robot that it will not proceed toward a position that contains an obstacle. We also assume that if the human sees the robot moving to a workspace position, she will not intentionally move to block the robot's desired workspace position: 

\begin{equation}\label{eqn:d1}
\Phi_{d1} \triangleq \bigcirc RS = j \rightarrow \bigcirc \lnot \mathcal{O}_j \quad \forall j = \text{1,2,..N-1}.
\end{equation}

In our implementation, if there is a human occupying one of the robot's neighboring positions, the robot will ask her not to block the workspace. Thus, we assume that the person will move out of the way by the next time step: 

\begin{equation}\label{eqn:d2}
\begin{aligned}
    \Phi_{d2} \triangleq \mathcal{O}_j  \rightarrow \bigcirc \lnot \mathcal{O}_j \quad \forall j = \text{1,2,..N-1}.
\end{aligned}
\end{equation}

We also account for the possibility that the robot may not successfully drop off completed work in the Inventory Station on its first try. At all robot states, the robot either is or is not manipulating completed work (``hand full" $\triangleq HF$ is true or false). When $RS = 0$, if the robot is currently manipulating completed work, it will be successful ($\mathcal{S}$) or unsuccessful ($\lnot \mathcal{S}$) at dropping off the completed work in that time step. In order to prevent the environment from interfering indefinitely with a successful dropoff, we assume that no more than two consecutive tries are required to be successful. The associated specifications are written as follows: 

\begin{fleqn}
\begin{flalign}\label{eqn:d3}
    \Phi_{d3} \triangleq & (\lnot(RS = 0) \wedge \bigcirc RS = 0 \wedge HF) \rightarrow &&\\\nonumber
    & \text{\qquad \qquad \qquad \quad} \bigcirc (HF \wedge \text{tries} = 1 \wedge (\lnot \mathcal{S} \vee \mathcal{S})),  
\end{flalign}
\end{fleqn}




\begin{flalign}\label{eqn:d4}
    \Phi_{d4} \triangleq & (RS = 0 \wedge HF \wedge \lnot \mathcal{S} \text{$\wedge$ tries = 1}) \rightarrow &&\\\nonumber
    & \text{\qquad \qquad \qquad \quad} \bigcirc (HF \wedge \text{tries} = 2 \wedge \mathcal{S}), \: \text{and}
\end{flalign}



\newcommand{\spacefix}{$\:$ }
\begin{fleqn}

\begin{flalign}\label{eqn:d5}
     \Phi_{d5} \triangleq (RS = 0 \wedge HF \wedge \mathcal{S}) \rightarrow \bigcirc (\lnot HF).
\end{flalign}
\end{fleqn}

As discussed in Sec. \ref{sec:delivery_human_model}, when the robot is not dropping off deliverables, there is uncertainty in how much $BL$ will decrease as the robot transition between states in the real environment. This uncertainty will impact the value of $BL$ at the next time step when the human is working. To express the specifications for $BL$ reduction, we define a formula that describes the situations in which the robot will attempt to drop off completed work:

\begin{equation}
g_4 \triangleq \big[\{\lnot(RS = 0) \wedge \bigcirc RS = 0\} \vee (\text{tries} = 1 \wedge \lnot \mathcal{S})\big]  \wedge HF. 
\end{equation}



We now distinguish between the robot moving within the workspace and performing the dropoff behavior. For workspace motions, we define

\begin{equation}\label{eqn:d6}
    \Phi_{d6} \triangleq  \lnot (\bigcirc RS = \text{N}) \wedge \lnot g_4 \wedge \lnot \text{wait} \rightarrow \bigcirc v_1,
\end{equation}

\noindent where $v_1$ is as defined in (\ref{eqn:wl_small_decrease}). During dropoff at the Inventory Station the following specification holds: 

\begin{equation}\label{eqn:d7}
    \Phi_{d7} \triangleq g_4 \wedge \lnot \text{wait} \rightarrow \bigcirc v_2,
\end{equation}

\noindent where $v_2$ is as defined in (\ref{eqn:wl_big_decrease}). We desire that the robot always eventually drops off completed work at the Inventory Station.  Unless $BL$ decreases, there would be no guarantee that the robot would always eventually have completed work to pick up from the human and drop off at the Inverntory Station. We add the assumption that if $BL$ stays constant in two consecutive time steps, then $BL$ will decrease in the next time step:

\begin{equation}\label{eqn:d8}
\begin{aligned}
    \Phi_{d8} \triangleq \lnot (\bigcirc RS = \text{N}) \wedge v_4 \wedge \lnot \text{wait} \rightarrow & \bigcirc (\bigvee_{k=1}^{5} (BL-k\gamma)), 
\end{aligned}
\end{equation}


\noindent where $v_4$ is as defined in (\ref{eqn:BL_stays_constant}). We synchronize the real robot's motion with the $BL$ model so that this assumption is valid in our implementation.

\subsection{Controller Synthesis}
The reactive synthesis problem was implemented using Slugs \cite{ehlers2016slugs}  with N = 3. (We consider four states for our proof-of-concept hardware implementation in Sec. \ref{sec:experiments}, but the underlying techniques of our approach can handle a much larger number of states.) By formulating the problem as a two-player game, Slugs can construct a reactive robot controller that upholds our specification of interest within $TS_{HRI}$.

In the simulation, the robot starts at State 0, $\gamma = 3.3\%$, and $\delta = 50\%$. In order to strike a balance between state space fineness and computational efficiency, $BL$ is represented in Slugs as 0,1,2,...,30, which corresponds to 0$\%$,3.3$\%$,6.7$\%$,...,100$\%$. Slugs was used to synthesize a controller that always satisfies the specification that the human's backlog never reaches 0$\%$ and never exceeds 87$\%$. In this manner, the human always has work to complete, but the robot does not seek to stress her. It is also desired that the robot will return to the Inventory Station infinitely often to drop off completed work. Since $BL$ can vary from 0 to 30 in Slugs, we express the system specification as: 

\begin{equation}\label{eqn:specfication}
    \Phi_1 = (\square BL \leq 26) \wedge (\square BL > 0) \wedge \square \lozenge (RS = 0 \wedge  HF)
    \end{equation}

We provide as an initial condition that $BL_{init}$ is between 30$\%$ and 86.7$\%$ ($9 \leq BL_{init} \leq 26$), as we found that outside of this $BL_{init}$ range, (\ref{eqn:specfication}) is not realizable. We synthesize a strategy using a quad-core Intel Core i7 processor and 12GB of RAM. Slugs computes in less than five seconds that the specification is realizable, and devises a high-level controller that guarantees that the robot will react properly to its environment while upholding the system specification. The controller is in the form of a decision tree, with nodes that capture all possible combinations of robot and environment behaviors, and the possible transitions from each node. Based on the robot's present state and the environment's behavior, the decision tree determines the appropriate next robot action. We now have a policy that can be leveraged for online decision-making. 

\section{Experiments}\label{sec:experiments}

\subsection{Toyota Human Support Robot}

 The Toyota Human Support Robot (HSR), which comprises an omni-directional base and a 5-DOF single manipulator, was adopted as the hardware platform for experimentation. The HSR uses two different computers: The main PC is for primary perception, navigation, and manipulation tasks. An Alienware laptop (Intel Core i7-7820HK, GTX 1080) is used for running OpenPose\footnote{https://github.com/CMU-Perceptual-Computing-Lab/openpose}, a real-time convolution neural network based algorithm used for human detection. All robot sub-programs communicate with each other via the Robot Operating System (ROS) interface. An overview of the HSR skills used for controller implementation is provided below.  

\subsubsection{Perception}
A laser range scanner is used on both sides of the mobile base to detect whether there is an obstacle within approximately one meter of the base. To simulate a more realistic working environment, the robot also perceives if there is a human in the workstation or if she has left to take a break. A depth camera for RGB-D video streaming is located on the HSR's head.  Recognition of whether or not there is a human in the workstation, based on the RBG-D data, is executed by OpenPose. We created a ROS action so that the HSR turns its head toward the workstation every five seconds to check for worker presence. 

\subsubsection{Navigation}
All basic navigation functions, including wheel-joint control and avoiding obstacles,  are  included  in the ROS navigation stack. It is assumed that given a goal position, the robot can safely navigate to this location, while avoiding dynamic obstacles via a re-planning scheme.

\subsubsection{Manipulation}
The robot's manipulator is used to pick up completed work from the human, and to drop off completed work at the Inventory Station. For pickup, the robot moves its end effector near to the human's right hand so that the human can hand over her work. To distinguish between $\mathcal{S}$ and $\lnot \mathcal{S}$ during dropoff, as discussed in Sec. \ref{sec:reactive_synthesis}, we use a force sensor mounted at the end effector to judge whether or not the object successfully made contact with the counter.

\begin{figure}[tp]
\def\svgwidth{.95\columnwidth}
\begin{center}
  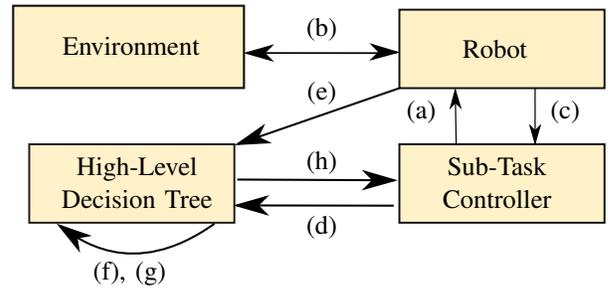
    \vspace{-12pt}  
  \caption{The communication protocols between the the robot and the sub-task (SMACH) and high-level (Slugs) controllers. The systems communicate by repeating (a)-(h), where: (a) Track and command 
sequential tasks; (b) Perceive, navigate, and manipulate; (c) Sequential tasks complete; (d) Request next action; (e) Subscribe to ROS environment topics; (f) Check for human at workstation. Update $RS$ and $BL$ when worker is present; 
 (g) Select next action from 
look-up table.  
(h) Respond with next action.}  \label{fig:architecture}
    \vspace{-22pt}  
\end{center}
\end{figure}

\subsection{Controller Implementation} 
The decision tree produced by Slugs serves as an online look-up table during robot operation. After transitioning to the next commanded state, the HSR will update its knowledge of the environment (mainly, any obstacles, if a dropoff action was successful, and the human's current $BL$). We used SMACH\footnote{http://wiki.ros.org/smach} to implement a finite-state machine framework that bridges the gap between the high-level action policy from Slugs and the robot's lower-level sequential task executors. 

Once the Slugs planner determines the next desired action, the robot's required skills are executed sequentially by the sub-task controller that is associated with a particular SMACH state. In other words, the sub-task control layer is responsible for decomposing the desired Slugs action into the sequential, lower-level required skills. For example, when the robot has completed work, the action $Go_{S0}$ first contains navigation (move to counter at State 0), then manipulation (drop off object), and finally navigation (move back from counter)  skills. The sequence of behaviors thus requires recognition of when the previous sub-task succeeds. All robot sub-tasks are programmed with the structure of ROS actions in order to be flexible to sub-task execution times. Once a high-level action is completed, or the first dropoff try is unsuccessful, SMACH requests the next desired action from the Slugs planner. The system architecture is shown in Fig. \ref{fig:architecture}.  

\begin{figure}[tp]
\def\svgwidth{.95\columnwidth}
\begin{center}
\begingroup%
  \makeatletter%
  \providecommand\color[2][]{%
    \errmessage{(Inkscape) Color is used for the text in Inkscape, but the package 'color.sty' is not loaded}%
    \renewcommand\color[2][]{}%
  }%
  \providecommand\transparent[1]{%
    \errmessage{(Inkscape) Transparency is used (non-zero) for the text in Inkscape, but the package 'transparent.sty' is not loaded}%
    \renewcommand\transparent[1]{}%
  }%
  \providecommand\rotatebox[2]{#2}%
  \newcommand*\fsize{\dimexpr\f@size pt\relax}%
  \newcommand*\lineheight[1]{\fontsize{\fsize}{#1\fsize}\selectfont}%
  \ifx\svgwidth\undefined%
    \setlength{\unitlength}{56.25bp}%
    \ifx\svgscale\undefined%
      \relax%
    \else%
      \setlength{\unitlength}{\unitlength * \real{\svgscale}}%
    \fi%
  \else%
    \setlength{\unitlength}{\svgwidth}%
  \fi%
  \global\let\svgwidth\undefined%
  \global\let\svgscale\undefined%
  \makeatother%
  \begin{picture}(1,0.53333333)%
    \lineheight{1}%
    \setlength\tabcolsep{0pt}%
    \put(0,0){\includegraphics[width=\unitlength,page=1]{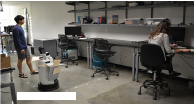}}%
    \put(0.23359447,0.02905467){\color[rgb]{0,0,0}\makebox(0,0)[t]{\lineheight{1.25}\smash{\begin{tabular}[t]{c}Robot at State 2\end{tabular}}}}%
    \put(0,0){\includegraphics[width=\unitlength,page=2]{workspace.pdf}}%
    \put(0.29224895,0.45556194){\color[rgb]{0,0,0}\makebox(0,0)[t]{\lineheight{1.25}\smash{\begin{tabular}[t]{c}Obstacle at State1\end{tabular}}}}%
    \put(0,0){\includegraphics[width=\unitlength,page=3]{workspace.pdf}}%
    \put(0.23945164,0.40072924){\color[rgb]{0,0,0}\makebox(0,0)[lt]{\lineheight{1.25}\smash{\begin{tabular}[t]{l}Inventory (State 0)\end{tabular}}}}%
    \put(0,0){\includegraphics[width=\unitlength,page=4]{workspace.pdf}}%
    \put(0.56710935,0.01338443){\color[rgb]{0,0,0}\makebox(0,0)[lt]{\lineheight{1.25}\smash{\begin{tabular}[t]{l}Workstation (State 3)\end{tabular}}}}%
  \end{picture}%
\endgroup%

  \caption{Experimental setup with four positions in the workspace, corresponding to $RS$ = 0 (Inventory Station), 1, 2, and 3 (Human Workstation). The HSR transports deliverables in its satchel and carries completed work in its manipulator back to the Inventory Station. A human obstructs its path, causing it to remain in State 2 until the next time step, by which time the human will have departed.}  \label{fig:workspace}
    \vspace{-20pt}  
\end{center}
\end{figure}

\subsection{Results}

Through experimentation, we sought to verify that our automatically-synthesized, high-level controller properly reacts to its environment while maintaining (\ref{eqn:specfication}). Fig. \ref{fig:workspace} shows the layout of our experimental setup and the positions of States 0 through 3. Referring to the $BL$ model in (\ref{eqn:BL_model}), we take $t_d = 10 s$.

\begin{figure}[tp]
\def\svgwidth{.9\columnwidth}
\begin{center}
  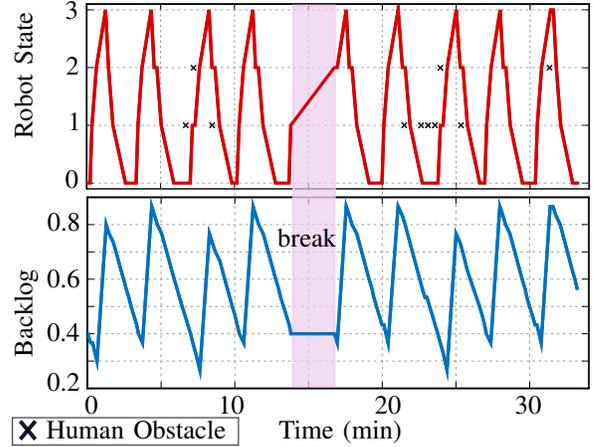
  \caption{Robustness test with $BL_{init}$ = 40$\%$. Human obstacles that appear in States 1 and 2 and depart by the next time step are marked by black x's. The worker moves out of the workspace for three minutes, which is highlighted in pink. The robot waits for the human to return at $RS = 2$, but does not update $RS$ and the environment variables in the Slugs planner until the human returns to work.}  \label{fig:30_min}
    \vspace{-22pt}  
\end{center}
\end{figure}

To test the robustness of the controller, we allowed the HSR  to operate autonomously for 30 minutes. During this time, the robot reacted to obstacles, interacted with the worker at the workstation, and returned to the Inventory Station several times to drop off completed work, as shown in Fig \ref{fig:30_min}. While we did not account for the human taking a break in our reactive synthesis problem, we incorporated this consideration for our experiments. If the HSR does not sense a human at the workstation, the HSR waits until she reappears, and then proceeds with its actions according to the Slugs planner.  

\begin{figure}[tp]
\def\svgwidth{.98\columnwidth}
\begin{center}
  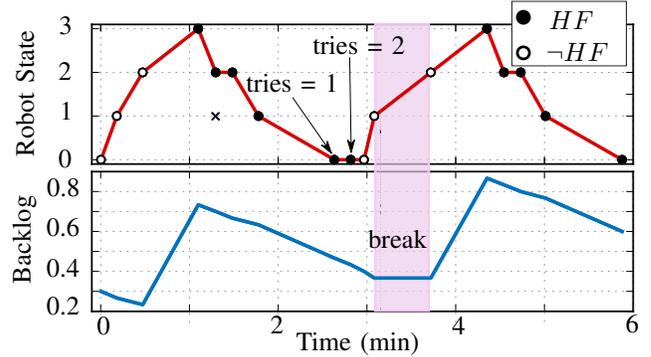
  \caption{Robot behavior with $BL_{init}$ = 30$\%$. When $RS = 1$ and $HF$ at 1.75 minutes, the SMACH sub-task controller tracks the amount of time the robot takes to travel from State 1 to State 0 and execute the dropoff sub-tasks, during which time, tries = 1. If this time exceeds 35 seconds, $\lnot \mathcal{S}$ is communicated to the high-level controller. The Slugs planner updates the tries value to be equal to 2, and commands that the robot continue to execute its dropoff sub-tasks in the next time step. After the second try, the robot successfully drops off the work, and the robot's manipulator is empty again.}  \label{fig:hand_full}
    \vspace{-18pt}  
\end{center}
\end{figure} 

\begin{figure}[tp]
\def\svgwidth{.98\columnwidth}
\begin{center}
  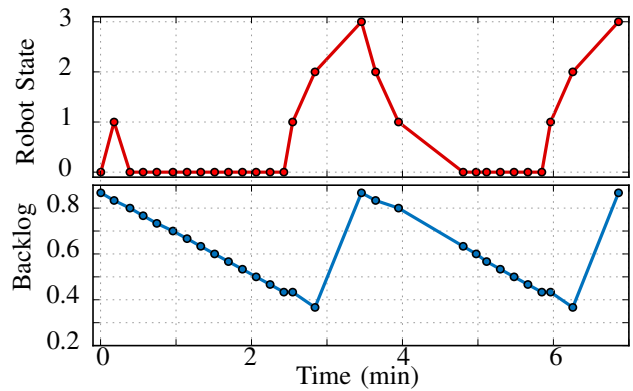
  \caption{Robot behavior with $BL_{init}$ = 86.7$\%$. The circular markers indicate the times at which  the sub-task controller calls the high-level decision tree to request the next robot action. Excluding its initial movement to and from State 1, the robot elects to wait at State 0 until it moves to deliver work, even when there are no obstacles present.}  \label{fig:init_86}
    \vspace{-18pt}  
\end{center}
\end{figure} 

It is also of interest to investigate the high-level controller behavior for differing $BL_{init}$ values. Fig. \ref{fig:hand_full} shows the results with $BL_{init}$ = 30$\%$ ($BL_{init} = \frac{9}{30}$ in Slugs).\footnote{This experiment is shown in the included video, which is also available at https://youtu.be/My6WIiZZCsM.} We also highlight whether the robot is manipulating completed work, and the implementation of the work dropoff logic at State 0. Fig. \ref{fig:init_86} shows the robot's behavior with $BL_{init}$ = 86.7$\%$ ($BL_{init} = \frac{26}{30}$ in Slugs). We note that the robot first moves to State 1, moves back to State 0, and remains at this position until it proceeds to the workstation to provide deliverables. The initial movement to and from State 1 upholds a winning strategy in the two-player game, but provides no useful output. Additional specifications could be imposed in the reactive synthesis problem to incorporate robot energy efficiency as a consideration to eliminate unnecessary robot movements. 

In all three experiments, the robot successfully maintains human $BL$ needs and can autonomously reason about and react to its environment. In the situations considered, the $BL$ value reaches 86.7$\%$, but does not exceed 87$\%$, as desired. It is interesting to note that the controller strategies drive the robot to wait in State 0 until it moves to deliver new work. The robot does not wait in States 1 or 2 for extended periods of time. This behavior is sufficient to satisfy system specifications, but optimizing where the robot waits could be needed in other work environments. For the experimental setup considered, we have verified that our proposed controller offers robustness and flexibility for a real-time, human-centered application. 

\section{Discussion}
Human factors are often neglected in formal methods when devising robot controllers, at the expense of having no formal guarantees that the robot can actually realize a human-centered goal. In this paper, we leverage reactive synthesis to devise a controller that supports a human's and a robot's collaborative goal. By formulating  a work delivery scenario as a two-player game, we automatically synthesize a controller that considers a human's work needs and supports robustness to system disturbances. Experimental validation on the HSR hardware demonstrates that the high-level controller strategy enables the robot to operate autonomously and robustly, while considering the activities and productivity of the human worker.

There are several interesting areas remaining for future work. The human backlog model in this case study is a toy model used for proof-of-concept. In the future, we are interested to incorporate verified, dynamic human models based on psychology and cognitive theory, to consider how factors such as fatigue, training, motivation, and stress impact backlog. We are also interested to study how we may extend our work to produce not only feasible motions, but also optimal motions. For now, our results support the feasibility of designing robots that are formally guaranteed to reduce the human share of work activities in the execution of everyday tasks.



\bibliographystyle{bib/IEEEtran}
\bibliography{bib/bib}

\begin{thebibliography}{10}
\providecommand{\url}[1]{#1}
\csname url@samestyle\endcsname
\providecommand{\newblock}{\relax}
\providecommand{\bibinfo}[2]{#2}
\providecommand{\BIBentrySTDinterwordspacing}{\spaceskip=0pt\relax}
\providecommand{\BIBentryALTinterwordstretchfactor}{4}
\providecommand{\BIBentryALTinterwordspacing}{\spaceskip=\fontdimen2\font plus
\BIBentryALTinterwordstretchfactor\fontdimen3\font minus
  \fontdimen4\font\relax}
\providecommand{\BIBforeignlanguage}[2]{{%
\expandafter\ifx\csname l@#1\endcsname\relax
\typeout{** WARNING: IEEEtran.bst: No hyphenation pattern has been}%
\typeout{** loaded for the language `#1'. Using the pattern for}%
\typeout{** the default language instead.}%
\else
\language=\csname l@#1\endcsname
\fi
#2}}
\providecommand{\BIBdecl}{\relax}
\BIBdecl

\bibitem{sutherland2007seven}
J.~Sutherland and B.~Bennett, ``The seven deadly wastes of logistics: applying
  toyota production system principles to create logistics value,'' \emph{White
  paper}, vol. 701, pp. 40--50, 2007.

\bibitem{kruse2013human}
T.~Kruse, A.~K. Pandey, R.~Alami, and A.~Kirsch, ``Human-aware robot
  navigation: A survey,'' \emph{Robotics and Autonomous Systems}, vol.~61,
  no.~12, pp. 1726--1743, 2013.

\bibitem{maniatopoulos2016reactive}
S.~Maniatopoulos, P.~Schillinger, V.~Pong, D.~C. Conner, and H.~Kress-Gazit,
  ``Reactive high-level behavior synthesis for an atlas humanoid robot,'' in
  \emph{Robotics and Automation (ICRA), 2016 IEEE International Conference
  on}.\hskip 1em plus 0.5em minus 0.4em\relax IEEE, 2016, pp. 4192--4199.

\bibitem{he2017reactive}
K.~He, M.~Lahijanian, L.~E. Kavraki, and M.~Y. Vardi, ``Reactive synthesis for
  finite tasks under resource constraints,'' in \emph{Intelligent Robots and
  Systems (IROS), 2017 IEEE/RSJ International Conference on}.\hskip 1em plus
  0.5em minus 0.4em\relax IEEE, 2017, pp. 5326--5332.

\bibitem{zhao2016high}
Y.~Zhao, U.~Topcu, and L.~Sentis, ``High-level planner synthesis for whole-body
  locomotion in unstructured environments,'' in \emph{Decision and Control
  (CDC), 2016 IEEE 55th Conference on}.\hskip 1em plus 0.5em minus 0.4em\relax
  IEEE, 2016, pp. 6557--6564.

\bibitem{wang2016task}
Y.~Wang, N.~T. Dantam, S.~Chaudhuri, and L.~E. Kavraki, ``Task and motion
  policy synthesis as liveness games,'' in \emph{ICAPS}, 2016, p. 536.

\bibitem{ramachandran2017give}
A.~Ramachandran, C.-M. Huang, and B.~Scassellati, ``Give me a break!:
  Personalized timing strategies to promote learning in robot-child tutoring,''
  in \emph{Proceedings of the 2017 ACM/IEEE International Conference on
  Human-Robot Interaction}.\hskip 1em plus 0.5em minus 0.4em\relax ACM, 2017,
  pp. 146--155.

\bibitem{chen2018planning}
M.~Chen, S.~Nikolaidis, H.~Soh, D.~Hsu, and S.~Srinivasa, ``Planning with trust
  for human-robot collaboration,'' in \emph{Proceedings of the 2018 ACM/IEEE
  International Conference on Human-Robot Interaction}.\hskip 1em plus 0.5em
  minus 0.4em\relax ACM, 2018, pp. 307--315.

\bibitem{wu2017toward}
B.~Wu, B.~Hu, and H.~Lin, ``Toward efficient manufacturing systems: A trust
  based human robot collaboration,'' in \emph{2017 American Control Conference
  (ACC)}.\hskip 1em plus 0.5em minus 0.4em\relax IEEE, 2017, pp. 1536--1541.

\bibitem{webster2016toward}
M.~Webster, C.~Dixon, M.~Fisher, M.~Salem, J.~Saunders, K.~L. Koay,
  K.~Dautenhahn, and J.~Saez-Pons, ``Toward reliable autonomous robotic
  assistants through formal verification: a case study,'' \emph{IEEE
  Transactions on Human-Machine Systems}, vol.~46, no.~2, pp. 186--196, 2016.

\bibitem{kwiatkowska2017cognitive}
M.~Kwiatkowska, ``Cognitive reasoning and trust in human-robot interactions,''
  in \emph{International Conference on Theory and Applications of Models of
  Computation}.\hskip 1em plus 0.5em minus 0.4em\relax Springer, 2017, pp.
  3--11.

\bibitem{porfirio2018authoring}
D.~Porfirio, A.~Saupp{\'e}, A.~Albarghouthi, and B.~Mutlu, ``Authoring and
  verifying human-robot interactions,'' in \emph{The 31st Annual ACM Symposium
  on User Interface Software and Technology}.\hskip 1em plus 0.5em minus
  0.4em\relax ACM, 2018, pp. 75--86.

\bibitem{sadigh2016information}
D.~Sadigh, S.~S. Sastry, S.~A. Seshia, and A.~Dragan, ``Information gathering
  actions over human internal state,'' in \emph{2016 IEEE/RSJ International
  Conference on Intelligent Robots and Systems (IROS)}.\hskip 1em plus 0.5em
  minus 0.4em\relax IEEE, 2016, pp. 66--73.

\bibitem{rausch2018modeling}
M.~Rausch, A.~Fawaz, K.~Keefe, and W.~H. Sanders, ``Modeling humans: A general
  agent model for the evaluation of security,'' in \emph{International
  Conference on Quantitative Evaluation of Systems}.\hskip 1em plus 0.5em minus
  0.4em\relax Springer, 2018, pp. 373--388.

\bibitem{feng2016synthesis}
L.~Feng, C.~Wiltsche, L.~Humphrey, and U.~Topcu, ``Synthesis of
  human-in-the-loop control protocols for autonomous systems,'' \emph{IEEE
  Transactions on Automation Science and Engineering}, vol.~13, no.~2, pp.
  450--462, 2016.

\bibitem{baier2008principles}
C.~Baier and J.-P. Katoen, \emph{Principles of model checking}.\hskip 1em plus
  0.5em minus 0.4em\relax MIT press, 2008.

\bibitem{jorgensen2017exploring}
S.~J. Jorgensen, O.~Campbell, T.~Llado, D.~Kim, J.~Ahn, and L.~Sentis,
  ``Exploring model predictive control to generate optimal control policies for
  hri dynamical systems,'' \emph{arXiv preprint arXiv:1701.03839}, 2017.

\bibitem{heard2018survey}
J.~Heard, C.~E. Harriott, and J.~A. Adams, ``A survey of workload assessment
  algorithms,'' \emph{IEEE Transactions on Human-Machine Systems}, no.~99, pp.
  1--18, 2018.

\bibitem{macdonald2003impact}
W.~MacDonald, ``The impact of job demands and workload on stress and fatigue,''
  \emph{Australian Psychologist}, vol.~38, no.~2, pp. 102--117, 2003.

\bibitem{piterman2006synthesis}
N.~Piterman, A.~Pnueli, and Y.~Sa’ar, ``Synthesis of reactive (1) designs,''
  in \emph{International Workshop on Verification, Model Checking, and Abstract
  Interpretation}.\hskip 1em plus 0.5em minus 0.4em\relax Springer, 2006, pp.
  364--380.

\bibitem{ehlers2016slugs}
R.~Ehlers and V.~Raman, ``Slugs: Extensible gr (1) synthesis,'' in
  \emph{International Conference on Computer Aided Verification}.\hskip 1em
  plus 0.5em minus 0.4em\relax Springer, 2016, pp. 333--339.

\end{thebibliography}

\end{document}